\documentclass[runningheads]{llncs}
\usepackage{indentfirst} % 首行缩进
\usepackage{graphicx}
\usepackage{amsmath,amssymb} % define this before the line numbering.
\usepackage{color}
\usepackage{booktabs}
\usepackage{multirow}

\begin{document}
% \linenumbers
\pagestyle{headings}
\mainmatter
\def\ECCVSubNumber{7467}  % Insert your submission number here

%
%\titlerunning{ECCV-20 submission ID \ECCV20SubNumber}
%
%\authorrunning{ECCV-20 submission ID \ECCV20SubNumber}
%
%%\author{Anonymous ECCV submission}
%\author{Lele Cheng, Xiangzeng Zhou, Liming Zhao, Dangwei Li, \\Hong Shang, Yun Zheng, Pan Pan, Xu Yinghui}
%\institute{Alibaba DAMO Academy}
%%\email{\{yinan.cll,zengyang.zxz\}@alibaba-inc.com}

\title{Weakly Supervised Learning with Side Information for Noisy Labeled Images} % Replace with your title
%\titlerunning{ECCV-20 submission ID \ECCV18SubNumber}
\titlerunning{Weakly Supervised Learning with Side Information for Noisy Labeled Images}

\author{Lele Cheng, Xiangzeng Zhou, Liming Zhao, Dangwei Li, \\Hong Shang, Yun Zheng, Pan Pan, Yinghui Xu}
%\authorrunning{ECCV-20 submission ID \ECCV18SubNumber}
\authorrunning{L. Cheng et al.}

\institute{Machine Intelligence Technology Lab, Damo Academy, Alibaba Group, China\\
\email{\{yinan.cll,xiangzeng.zxz,lingchen.zlm,dangwei.ldw,shanghong.sh,\\zhengyun.zy,panpan.pp\}@alibaba-inc.com,renji.xyh@taobao.com}}

\maketitle

\begin{abstract}
In many real-world datasets, like WebVision, the performance of DNN based classifier is often limited by the noisy labeled data.
To tackle this problem, some image related side information, such as captions and tags, often reveal underlying relationships across images.
In this paper, we present an efficient weakly-supervised learning by using a Side Information Network (SINet), which aims to effectively carry out a large scale classification with severely noisy labels. The proposed SINet consists of a visual prototype module and a noise weighting module. The visual prototype module is designed to generate a compact representation for each category by introducing the side information. The noise weighting module aims to estimate the correctness of each noisy image and produce a confidence score for image ranking during the training procedure. The propsed SINet can largely alleviate the negative impact of noisy image labels, and is beneficial to train a high performance CNN based classifier.
Besides, we released a fine-grained product dataset called AliProducts, which contains more than 2.5 million noisy web images crawled from the internet by using queries generated from 50,000 fine-grained semantic classes.
Extensive experiments on several popular benchmarks~(i.e. Webvision, ImageNet and Clothing-1M) and our proposed AliProducts achieve state-of-the-art performance. The SINet has won the first place in the 5000 category classification task on WebVision Challenge 2019, and outperforms other competitors by a large margin.

\keywords{weakly supervised learning, noisy labels, side information, large scale web images}
\end{abstract}

\section{Introduction}
In recent years, the computer vision community has witnessed the significant success of Deep Neural Networks (DNNs) on several benchmark datasets of image classification, such as ImageNet~\cite{imagenet} and MS-COCO~\cite{coco}.
However, obtaining large-scale data with clean and reliable labels is expensive and time-consuming. When noisy labels are introduced in training data, it is widely known that the performance of a deep model can be significantly degraded~\cite{ref1,cloth1m,ref2,zhu-review-2004}, which prevents deep models from being quickly employed in real-world noisy scenarios.

A common solution is to collect a large amount of image related side information~(e.g. surrounding texts, tags and descriptions) from the internet, and directly take them as the ground-truth for model training. Though this solution is more efficient than manual annotation, the obtained labels usually contain noise due to the heterogeneous sources. Therefore, improving the robustness of deep learning models against noisy labels has become a critical issue.

To estimate the noise in labels, some works propose new layers~\cite{Sukhbaatar-training-2014,Goldberger-training-2016} or loss functions~\cite{Patrini-making-2017,Hendrycks-using-2018,Zhang-Generalized-2018,ref20} to correct the noisy label during training. However, these works rely on a strict assumption that there is a single transition probability between the noisy labels and the ground-truth labels.
Owning to this assumption, these methods may show good performance on hand-crafted noisy datasets but are inefficient on real noisy datasets such as Clothing1M~\cite{cloth1m}.
%Although manually annotating all training samples is costly, i
In some situations, it is possible to annotate a small fraction of training samples as additional supervision. By using  additional supervision, works like~\cite{Li-learning-2017,ref12,Veit-learning-2017} could improve the robustness of deep networks against label noises.
But still, the requirement on clean samples make them less flexible to apply in large scale real-world scenarios.

Many data cleaning algorithms~\cite{Brodley-Identifying-2011,Miranda-use-2009,Barandela-Decontamination-2000} are developed to discard those samples with wrong label ahead of the training procedure. The major difficulty of these algorithms is how to distinguish informative hard samples from harmful mislabeled ones. CleanNet~\cite{ref12} achieves state-of-the-art performance on the real-world noisy dataset Clothing1M~\cite{cloth1m}. CleanNet generates a single representative sample~(class prototype) for each class and uses it to estimate the correctness of sample labels. With the observation that samples have wide-spread distribution in noisy classes, SMP~\cite{ref22} takes multiple prototypes to represent a noisy class instead of single prototype in CleanNet. In both CleanNet and SMP, extra clean supervision is required to train models.

In most of previous works, image related side information or annotations~(e.g. titles and tags) from web are commonly regarded as noisy labels. These works may not fully take advantage of the side information. Based on our observations, these image related side information reveal underlying similarity among images and classes, which has great potential to help tackle label noises.
%In this paper, the side information of image label structure and text descriptions as regarded as weakly annotations. Through analyzing the label structure and descriptions,
By analyzing the label structure and text descriptions, we explore an weakly-supervised learning strategy to deal with noisy samples. For example, the label ``apple'' may refer to a fruit or an Apple mobile phone. When acquiring images from web using the label ``apple'', images of apple fruits and Apple mobile phones will be wrongly put under a same class. Fortunately, titles or text descriptions about the images could imply the misplacement.
In this paper, we propose an efficient weakly-supervised learning strategy to evaluate the correctness of each image sample in each class by exploiting the label structure and label descriptions. Moreover, we release a large scale fine-grained product dataset to facilitate further research on visual recognition. To our knowledge, the proposed product dataset contains the largest number of product categories by now.
%Along with the released image data, image titles and class hierarchy are also provided for the convenience of extending research.

%Regarding the image related side information as weakly annotations, we propose an efficient unsupervised learning strategy, which evaluating the correctness of each image sample in each class by analyzing the label structure and label descriptions. A process of training images ranking is conducted with the generated correctness confidences. On the other hand, in order to facilitate further research on exploring visual recognition both academically and industrially, we release a large scale fine-grained product dataset. The proposed dataset collects more than $2.5$ million product images crawled from the Internet by using queries generated from $50000$ fine-grained semantic concepts. To our knowledge, the proposed product dataset contains the most number of product categories by now. Along with the released image data, image titles and class hierarchy are also provided for the convenience of extending research.

%pic1: webvision
\begin{figure}[htbp]
\centering
\includegraphics[height=5.8cm, width=12cm]{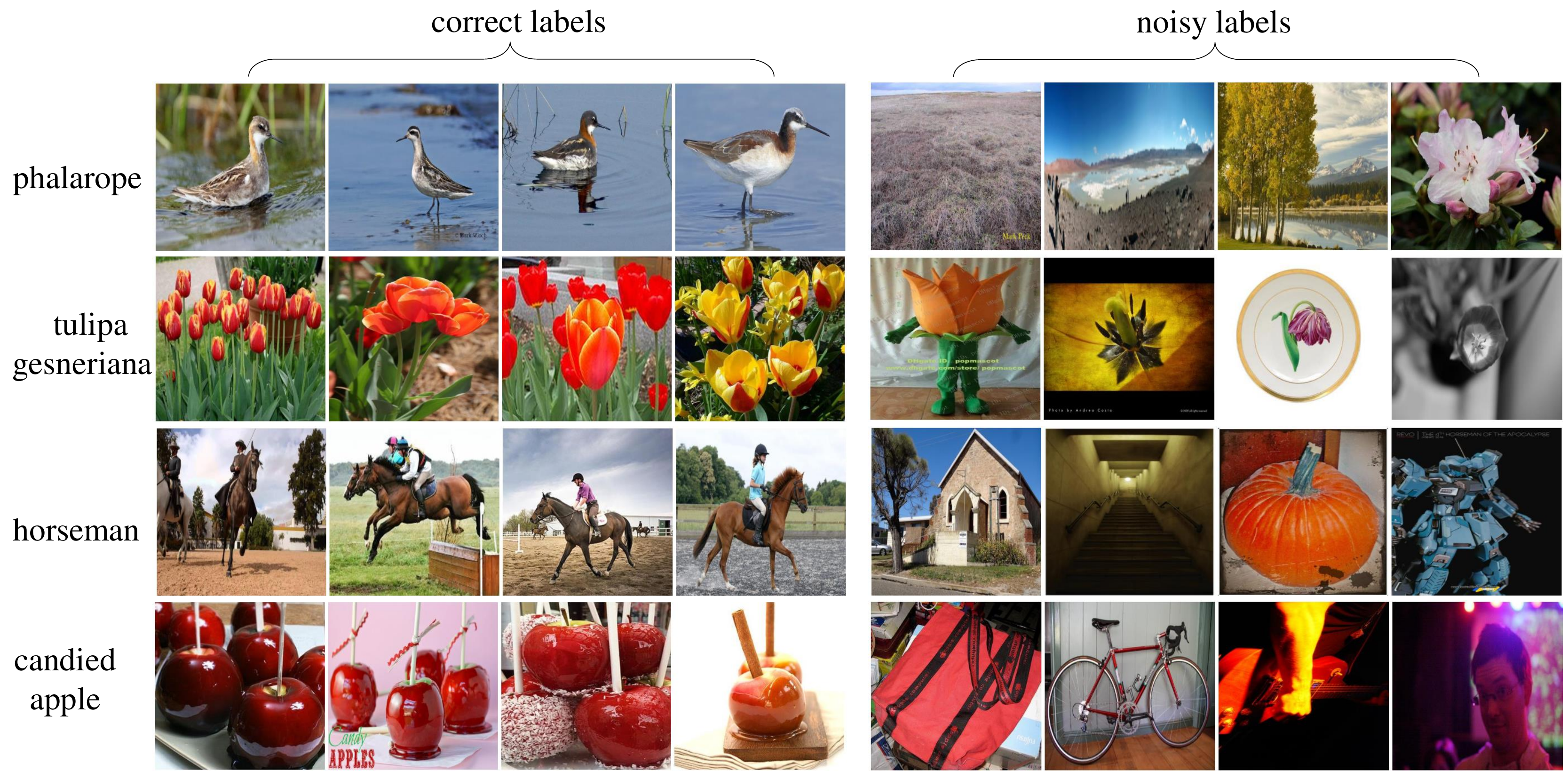}
\caption{Images of WebVision 2019 dataset~\cite{webvision} from the categories of phalarope, horseman, candied apple, tulipa, gesneriana
. The dataset was collected from the Internet by textual queries
generated from 5, 000 semantic concepts in WordNet. Obviously,
each category includes a lot of noisy images as shown above.}
%\vspace{-0.5cm}
\end{figure}

Our contributions in this paper are summarized as follows:

1) A weakly supervised learning with side information network~(SINet) is proposed for noisy labeled image classification. SINet infers the relationship between images and labels without any human annotation, and enable us to train high-performance and robust CNN models against large scale label noises.
% which enable us to train high-performance CNN models for large scale web noisy images by directly exploring visual and label relations without any human annotation.

2) A noisy and fine-grained product dataset called AliProducts is released, which contains more than $2.5$ million web images crawled from the Internet by using queries generated from the $50,000$ fine-grained semantic classes. In addition, side
information (e.g., hierarchical relationships between classes) are also provided for the convenience of extending research.

3) Extensive experiments are conducted on a number of benchmarks, including WebVision, ImageNet, Clothing1M and AliProducts, in which the proposed SINet obtains the state-of-the-art performance. Our SINet also won the first place on the WebVision Challenge 2019, and outperforms the other competitors by a large margin.

\section{Related Work}
Recent studies have shown that the performances of DNNs degraded substantially when training on data with noisy labels \cite{ref1,ref2}.
To alleviate this problem, a number of approaches have been introduced and can be generally summarized as below.

Some methods design robust loss functions against label noises \cite{ref3,ref4,ref5,ref6,ref7,ref10}.
Zhang et al.~\cite{ref4} found that the mean absolute error (MAE) is inherently more robust to label noises than the commonly-used categorical cross entropy (CCE) in many circumstances. However, MAE performs poorly with DNNs and
challenging datasets due to slow convergence. Generalized Cross Entropy
(GCE) loss \cite{ref10} applies a Box-Cox transformation to probabilities (power law function of probability with exponent
q) and can be seen as a generalization of MAE and CCE, thus can be easily applied with existing DNN architecture and yield good performance in certain noisy datasets.
%\textcolor{red}{Extending this work \cite{ref10} argues that MAE provides much smaller learning rate than categorical cross entropy (CCE), therefore a new loss function is suggested which combines robustness of MAE and implicit weighting of CCE. With a tuning parameter, it can be made more similar to MAE or CCE.}

Re-weighting training samples aims to evaluate the correctness of each sample on a given label, and has been widely studied in \cite{ref11,ref12,ref13,ref14,ref15,ref16,ref18,ref23}.
In \cite{ref13}, meta learning paradigm is used to determine the sample weighting factors.
\cite{ref14} takes open-set noisy labels into consideration and train a Siamese network to detect noisy labels.
In each iteration, sample weighting factors will be re-estimated, and the classifier will be updated at the same time.
\cite{ref15} also presents a method to separate clean samples from noisy samples in an iterative fashion.
The biggest challenge encountering these data cleaning algorithms is how to distinguish informative hard samples from harmful mislabeled ones.
To prevent discarding valuable hard samples, noisy samples is weighted according to their noisiness level which is estimated by pLOF \cite{ref16}.
In CleanNet~\cite{ref12}, an additional network is designed to decide whether a sample is mislabelled or not. CleanNet aims to produce weights of samples during the training procedure.
CurriculumNet~\cite{ref18} designs a learning curriculum by measuring the complexity of data and ranking samples in an unsupervised manner.
However, most of these approaches either requires extra clean samples as additional information or adopts a complicated training procedure, making them less suitable for being widely applied in many real-world scenarios.

Self-learning pseudo-labels has been studied in many scenarios to deal with noisy labels.
Reed et al.~\cite{ref19} propose to jointly train model with both noisy labels and pseudo-labels.
However, \cite{ref19} over-simplifies the assumption of the noisy label distribution, which leads to sub-optimal results.
In the joint optimization process of \cite{ref20}, original noisy labels are completely replaced by pseudo-labels. This often discards some valuable information in the original noisy labels. Li et al.~\cite{ref21} proposes to simulate actual training by generating synthetic noisy labels, and train the model such that after one gradient update using each set of synthetic noisy labels, thus the model does not overfit to the specific noise.

Our method is similar to the work of \cite{ref22}, in which each class is represented by a learnable prototype. For each sample, a similarity is calculated between the sample and the corresponding prototype to correct its label. A final classifier is trained by using both the corrected label and the noisy label. However, \cite{ref22} only takes visual information into consideration to construct class prototypes. Our approach integrate visual with side information to generate more reliable prototype for each class.

\begin{figure}[htbp]
\centering
\includegraphics[height=6.2cm, width=12cm]{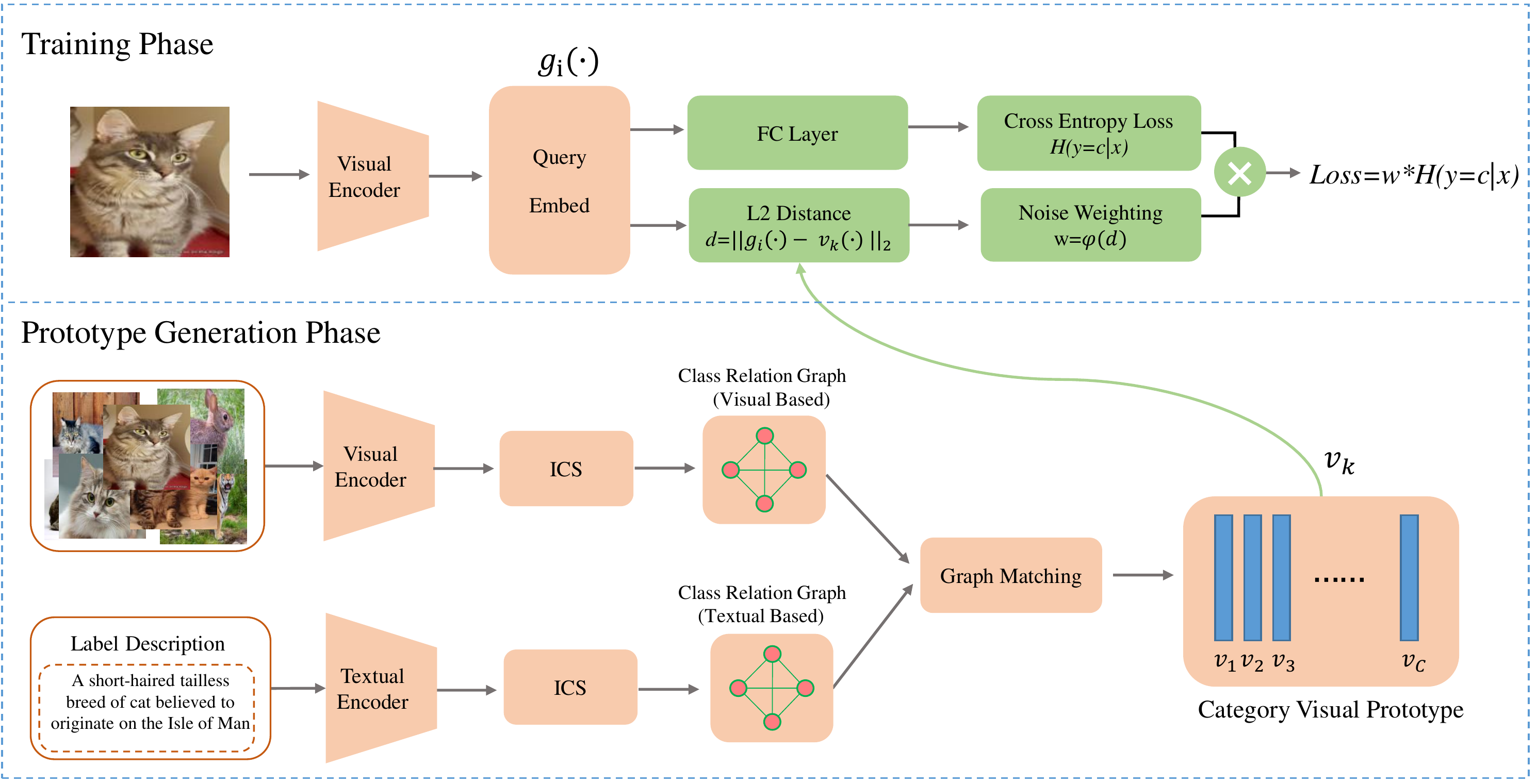}
\caption{Illustration of the framework of SINet on the noisy dataset, which includes three sub-modules, including \textit{Class Relation Graph},
\textit{Visual Prototype Generation}, and \textit{Noise Weighting}.
First, we construct visual-based and textual-based class relation graph with representation of image and textual using Inter Class Similarity (ICS), respectively.
Second, we  generate a visual prototype for each class using the compliance between the visual and textual class relations.
Finally, \textit{Noise Weighting} is used to weight all noisily labeled images with class prototypes before the training procedure.}
%\caption{Illustration of the framework of SINet on the noisy dataset, which includes three sub-modules \textit{Class Relation Graph},
%\textit{Visual Prototype Generation}, \textit{Noise Weighting}. First, we utilize image related side information to construct the textual based \textit{Class Relation Graph}. Second, we construct visual based class relation graph with CNN and estimate the compliance between the visual and textual class relation, so as to generate a visual prototype for each class. Finally, \textit{Noise Weighting} is used to weight all noisily labeled images with class prototypes before the training procedure.}
%\vspace{-0.5cm}
\end{figure}

\section{Approach}
We focus on learning a robust image classifier from large-scale noisy images
with side information. Let $\mathbb{D} = \{(x_1, y_1),..., (x_N, y_N)\}$ be a noisily labeled dataset of N
images. $y_i \in \{1, 2, ...,C\}$ is the noisy label corresponding to the image $x_i$, and $C$ is the number of classes in the dataset.

%Since the images are noisily labeled,
%they would impede the model training and make it over-fitting the noisy labels.
Training with noisily labeled images, deep neural networks may over-fit these noisy labels and perform poorly.
To alleviate this problem, we introduce a conceptually simple
but effective side information network~(SINet) for training against
noisy labels.
Based on the knowledge of ``different classes look different"~\cite{Brust-Not-2019}, we train the network under the constraint that the produced visual similarity between classes should have potential relevance with their natural semantic similarity.
The semantic similarity is derived from a class relation graph constructed with the image related side information such as image titles, long text descriptions and image tags.

For each class, a prototype $v_k$, $k \in \{1,2,..C\}$ is generated from reliable training samples whose visual similarity graph aligns well with the constructed Class Relation Graph.
Subsequently, we can decide whether an training sample is mislabeled or not by comparing its visual
representation with $v_k$ which is considered as a clean and reliable representation of $k$-th class.
During the training phrase, an image is recognized as a noisy sample or not in the light of the distance of the image feature and the class prototype. Instead of directly discarding noisy images by a predefined threshold, we assign each image a correctness weight in a noise weighting module.

Training a deep model $\mathcal{G}$ parameterized by $\theta$ on the dataset $\mathbb{D}$, the overall optimization objective is formulated as
\begin{align}
  \theta^{*} = argmin \sum_{i=1}^{N}w_i*L(y_i,\mathcal{G}(\theta,x_i))
\end{align}
where $L$ is a conventional cross entropy loss, and $w_i$ is the image weight generated by the noise weighting module.

In the following sections, we elaborate the proposed SINet that using image related side information to facilitate a classification task on noisy images. The SINet comprises three modules, i.e. class relation graph generation, visual prototype generation and noise weighting. As shown in Fig.2, an overview of the SINet architecture is illustrated. In Section 3.1, two kinds of category relation graphs are constructed using label embeddings and WordNet information, respectively. In Section 3.2, the KL divergence is used to estimated the compliance between the two kinds of graphs, so as to generate a visual prototype for each class. Given the class prototypes, a noise weighting module is presented to weight all noisily labeled images before the training procedure in Section 3.3.

\subsection{Class Relation Graph}~\label{sec:graph}
In some classification scenarios, for each class we can obtain both the long-text description of label and the hierarchical structure of class relationships using WordNet~\cite{WordNet}. Both the label descriptions and WordNet structure reveal rich semantic information across classes. In this section, we attempt to exploit the inter-class semantic knowledge by constructing two kinds of class relation graphs. In the graphs, each node represents a class and the edges between nodes are built using two different similarity metrics.

Firstly, a straightforward way to build a class relation (marked as $\mathbb{G}_w$) is using the tree structure of WordNet. In the graph $\mathbb{G}_w$, an edge of two class nodes is created using the distance of the shortest path in the WordNet tree. Here we represent the WordNet-based class relation graph $\mathbb{G}_w$ as a matrix $S_w \in R^{C \times C}$.

Secondly, we learn a label embedding for each class node with the text description of label from WordNet. For instance, a label description \textit{Manx cat: A short-haired tailless breed of cat believed to originate on the Isle of Man} provides rich semantic knowledge of the corresponding class. Moreover, these text descriptions reveal underlying relationship across classes from the perspective of natural language.
To obtain a semantic representation of each class, we use a BERT~\cite{BERT} language model to learn a sequence of word embeddings from text description of each class label. We then feed them into a bidirectional LSTM module to achieve a class label level embedding (called label embedding). Please note that the number of label text descriptions available for training is too small, so we use a pretrained BERT and freeze it in the training procedure, and only finetune the LSTM module. Meanwhile, the number of trainable parameters is significantly reduced.

We then build a graph $\mathbb{G}_l$ based on the label embeddings, in which the set of nodes is $\mathcal{V}=\{\mathbf{v}_1,\mathbf{v}_2,...,\mathbf{v}_C\}$, and $\mathbf{v}_i \in R^d$ represents the label embedding of the $i$-th class. We then calculate the cosine similarity between all pairs of label embeddings to build edges of the graph $\mathbb{G}_l$. For convenience, the graph $\mathbb{G}_l$ is formulated as a inter-class similarity~(ICS) matrix $S_l \in R^{C \times C}$ as below.
\begin{align}
  S_l^{ij} = \frac{\mathbf{v}_i^T \mathbf{v}_j}{\left|\left|\mathbf{v}_i\right|\right|_2\left|\left|\mathbf{v}_j\right|\right|_2}
\end{align}
%
%After extracting the label embeddings for every class, we obtain inter-class relations by computing embeddings similarity. Then a preliminary class relation graph is constructed by using the inter-class similarity (ICS).
%\begin{align}
%  S_{ij} = \frac{e_i^T e_j}{\left|\left|e_i\right|\right|_2\left|\left|e_j\right|\right|_2}
%\end{align}
Then $S_l^{ij}$ is regarded as a kind of similarity between two class embeddings $\mathbf{v}_i$ and $\mathbf{v}_j$. Larger $S_l^{ij}$ indicates higher similarity between the classes $i$ and $j$.
%
%\begin{align}
%  S_e = ICS(E) = \{\{s(e_i,e_1),s(e_i,e_2)..s(e_i,e_C)\}, i={1,2,..,C}\}
%\end{align}
%where $S_e \in R^{C \times C}$ represents the inter-class relations, $E=\{e_1,e_2,...,e_C\}$, $e_i \in R^d$ represents the label embedding for $C$ classes, $s$ represents similarity function.

Eventually, we blend the two class relation graphs $\mathbb{G}_w$ and $\mathbb{G}_l$ generated using two kinds of semantic knowledge, and obtain a hybrid graph $\mathbb{G}_t$, formulated as below.
\begin{align}
  S_{t} = S_l + S_w
\end{align}
%where $\lambda$ is a trade-off coefficient, which is set as 1.0 in this paper.

%we compute the mutual relations by embeddings similarity and thus obtain a preliminary class relation graph. Besides, to incorporate the prior knowledge, we integrate the preliminary class relation graph with WordNet to get the ultimate class relation graph.

%while the label embedding provides a semantic description of each specific class.
\subsection{Visual Prototype Generation}
This section introduce an effective visual prototype generation module for training robust CNNs with noisy images. The key idea of visual prototype module is to generate a clean visual prototype $v_k$, $k \in \{1,2,..C\}$ for each class.
The visual prototype $v_k$ can be interpreted as a reliable and effective representation of $k$-th class, and can be used to identify the reliability of all training data.

In order to generate visual prototype $v_k$, we need to obtain some reliable images from $k$-th class, and evaluate their contributions to $v_k$. Since noisy images is ubiquitous within each class, it is an intractable problem to directly collect reliable images. In this paper, we resort to the class relation graph $\mathbb{G}_t$ constructed in Section~\ref{sec:graph} to help this collection. Intuitively, the inter-class relation in visual representation space for reliable images should be closely related to that in class relation graph. For example, the $k$ nearest classes of \textit{siamese cat} in class relation graph are \textit{persian cat}, \textit{tiger cat}, \textit{manx cat}, etc. If the $k$ nearest classes of an image in visual representation space are also the same, then this image is probably reliable, and should contribute to the generation of visual prototype in a high confidence.

%then use the average feature as the dummy representatin of this class.

To be specific, we consider an image sample $x_i$ and its current labelled class $c$. The semantic similarity vector of class $c$ can be obtained from the the class relation graph $\mathbb{G}_t$, and is denoted as a vector $s_t^i$ of length $C$.
To compute the visual similarity vector between $x_i$ and all $C$ classes, it is required to generate an initial prototype for each class first. We first extract visual features from all images using the CNN model in the proposed SINet.
Then for each class, top-$k$ ranked image features according to their classification confidence score are averaged to generate initial class prototype.
Then the visual similarity vector of $x_i$ is computed as the cosine similarity score of the CNN feature $g_i$ with all initial class prototypes, which is denoted a vector $s_v^i$ of length $C$.
%At the same time, through the backbone feature of the base model, we can obtain the visual k-nearest neighbours of $x_i$ in the feature space, and is denoted a vector $s_v^i$.
Finally the consistence score $p_i$ of image sample $x_i$ is estimated based on KL divergence between $s_t^i$ and $s_v^i$.
\begin{align}
  p_i = \frac{1}{(KL\left(\psi(s_t^{i}), \psi(s_v^{i})\right) + \epsilon)^\gamma}
  %p_i = {(KL\left(s_t^{i}, s_v^{i}\right) + \epsilon)^{-\gamma}}
\end{align}
where $\psi$ is a normalize function, eg. $L$2 norm or softmax, $\gamma$ is used to control the "contrast" of two similarity vectors, and $\epsilon$ is a small positive constant to prevent the denominator going to zero.

Eventually, we generate a visual prototype $v_c$ for class $c$ by using the weighted sum of the image features, as formulated in Equ.(5):
\begin{align}
  v_c = \frac{\sum_{i=1}^{N} g_i p_i}{\sum_{i=1}^{N}p_i}
\label{vk_soft}
\end{align}
where the $g_i$ is the visual CNN feature of image $x_i$ in the base model.

As the training proceeds, the visual prototype for each class will be updated iteratively, then more reliable samples could contribute to train the CNN model better.
%\begin{align}
%  v_c  =  VisualPrototype(V) = E\left[v*p_{clean}(v)\right] \\
%= \frac{\sum_{i=1}^{N} v_ip_{clean}\left(v_i\right)}{\sum_{i=1}^{N}p_{clean}\left(v_i\right)}
%\end{align}

\subsection{Noise Weighting}
In this section, we use the class prototypes generated above to weight noisily labelled images before training. Considering an image $x_i$ and its current labelled class $c$, we estimate an importance weight $w_{i,c}$ by calculating the Euclidean distance of the visual feature $g_i$ of image $x_i$ and the prototype $v_c$. As formulated in Equ.(6), the importance weight $w_{i,c}$ is computed with two hyper-parameters $\alpha$ and $\beta$ to control the shift and contrast of different visual features.

%This section describes the noise weighting for each image. The noise weighting module aims at estimating a
%confidence score $s_i$ about whether $x_i$ is a correctly labeled
%sample. For example, a tiger image is mistakenly collected
%into the cat category as shown in Fig.2. To effectively avoid over-fitting the noisy labels, we introduce the Noise Weighting module, which assign low confident score for the tiger image, while high confident scores for other cat images.

%To achieve this goal, we use following method for confidence score estimation. We estimate the confidence score based on the Euclidean distance between image embedding and visual prototype.
\begin{align}
  w_{i,c} = max\{0, \left[\alpha - ||v_c - g_i||_2 \right]^\beta\}
\end{align}
%where $\alpha$ and $\beta$ are two parameters in the score function to control the shift
%and contrast of different embeddings.
We finally use a weighted cross entropy loss for model training as shown in Equ.(7).
\begin{align}
  \mathrm{Loss}_{ce} = \sum_{i=1}^{N}\sum_{c=1}^{C}w_{i,c} \cdot log(p_{i,c})
\end{align}
where $p_{i,c}$ is the softmax output of image $x_i$ on class $c$.
%\subsection{Implementation Details}
\subsection{Implementation Details and WebVision Challenge.}
%\large{Web Noisy Data} AAA
%\textbf{Web Noisy Data} dataset
%\subsubsection{Web Noisy Data.} dataset
%\subsubsection{Final Results in the WebVision Challenge.} dataset
\footnotesize\textbf{Implementation Details.}
The scale of WebVision dataset is significantly larger than
that of ImageNet, so it is necessary to take into account the computational cost when
conducting extensive experiments in evaluation. In our experiments, we use resnext-101 as our standard architecture. Specifically, the weights of model are optimized with mini-batch SGD and the batch size is set to 2,500. The learning rate starts from 0.1, and decayed by a factor of 10 at the epochs of 30, 60, 80, 90. The model is trained for 100 epochs in total. To reduce the risk of over-fitting, we
use diverse data augmentations, such as random cropping, mirror flip and autoaugment. In addition, we also use dropout with a ratio
of 0.25 after the global average pooling layer.

%\footnotesize\textbf{Adaptive Label Smoothing.}
\footnotesize\textbf{Topk Label Smoothing.}
Since there exists massive noise images in WebVision, if we directly utilize the one-hot target of ground truth to train CNN, it is inevitable to over-fit the noisy labels. To alleviate this problem, we proposed Adaptive Label Smoothing to assist the model training. Specifically, we select a small subset of high confidence images to train an initial model, and then we use the model to predict  probability distribution of rest images. We use the topk predictions and ground truth to construct a smoothing label, and use this smoothing label to train the model. The Adaptive Label Smoothing enhance the tolerance of noisy labels, leading to
about 0.2\% performance improvements on top-5 accuracy in WebVision challenge.

\footnotesize\textbf{Adaptive Spatial Resolution.}
There exists a lot of fine-grained categories in WebVision, which are hard to distinguish. Many studies have show that high-resolution images can improve the performance of fine-grained recognition.
Inspired by this, we first train an initial model with fixed image resolution of 224x224, and then finetune the model with large input resolutions, e.g. 256x256 and 312x312. Specifically, the adaptive average pooling is used before the classifier layer to keep the feature dimension unchanged. The large input resolutions enhance the tolerance of noisy labels, leading to
about 0.5\% performance improvements on top-5 accuracy in WebVision challenge.

%\setlength{\tabcolsep}{1.4pt}
%\setlength{\tabcolsep}{4pt}
%\begin{table}
%\begin{center}
%\caption{Summary of popular general, fine-grained and noisy image datasets.}
%\label{ exp_AliProducts}
%\begin{tabular}{c|c|c|c|c}
%%\toprule  %添加表格头部粗线
%\hline
%Dataset & classes & images & characteristic & institute \\
%\hline
%ImageNet & 1,000 & 1.2M &  general &  Stanford \\
%\hline
%WebVision1.0 & 1,000 & 2.4M & noisy,long tail &  ETH Zurich \\
%\hline
%WebVision2.0 & 5,000 & 16M & noisy, long tail &  ETH Zurich \\
%\hline
%iNat2017 & 5,000 & 0.6M & fine-grained,long tail & iNaturalist \\
%\hline
%%iMat-Product & 2,019 & 1.1M & fine-grained &  Malong \\
%%\hline
%Clothing1M & 14 & 1M & noisy &  CUHK \\
%\hline
%\textbf{AliProducts} & 50,000 & 3.0M & noisy, fine-grained, long tail  & Alibaba \\
%\hline
%
%\setlength{\tabcolsep}{1.4pt}
%\setlength{\tabcolsep}{4pt}
%\begin{table}
%\begin{center}
%\caption{Summary of popular general, fine-grained and noisy image datasets.}
%\label{ exp_AliProducts}
%\begin{tabular}{c|c|c|c}
%%\toprule  %添加表格头部粗线
%\hline
%Dataset & classes & images & characteristic \\
%\hline
%ImageNet & 1,000 & 1.2M &  general \\
%\hline
%WebVision1.0 & 1,000 & 2.4M & noisy,long tail \\
%\hline
%WebVision2.0 & 5,000 & 16M & noisy, long tail \\
%\hline
%iNat2017 & 5,000 & 0.6M & fine-grained,long tail \\
%\hline
%%iMat-Product & 2,019 & 1.1M & fine-grained \\
%%\hline
%Clothing1M & 14 & 1M & noisy \\
%\hline
%\textbf{AliProducts} & 50,000 & 3.0M & noisy, fine-grained, long tail \\
%\hline
%%\bottomrule  %添加表格头部粗线
%\end{tabular}
%\end{center}
%%\vspace{-1cm}
%\end{table}

\section{Experiments}
In this section, we mainly evaluate our SINet on four popular benchmarks for noisy-labeled visual
recognition, i.e., WebVision, ImageNet, Clothing1M and AliProducts. Particularly, we investigate the
learning capability on large-scale web images without any human annotation.

\subsection{Datasets.}
\footnotesize\textbf{WebVision 1.0} ~\cite{webvision} is a large scale image dataset for object recognition and classification. The images are crawled from web search engine, e.g., Flickr and Google, by using 1,000 semantic concepts from WordNet. In addition, side information, e.g., tag, title, description, are also attached, which can be used to help the model training. WebVision contains 1,000 semantic categories and 2.4 millions images without any human annotation. Nonetheless, for the research purpose, 50k images with human annotation are used as validation and test set, respectively. Due to overlap between classes, the evaluation measure of WebVision is top-5 recognition accuracy.

\footnotesize\textbf{WebVision 2.0}  is similar with WebVision 1.0~\cite{webvision}. It also contains images crawled from the Flickr website and Google search engine. The number of visual concepts was extended from 1,000 to 5,000, and the total number of training images reaches 16 million. The dataset contains massive noisy labels, as shown in Fig. 1. In addition, 290K images with human annotation are used as validation and test set, respectively. The evaluation measure is the same as WebVision 1.0.

\setlength{\tabcolsep}{1.4pt}
\begin{figure}[htbp]
\centering
\includegraphics[height=5.0cm, width=12cm]{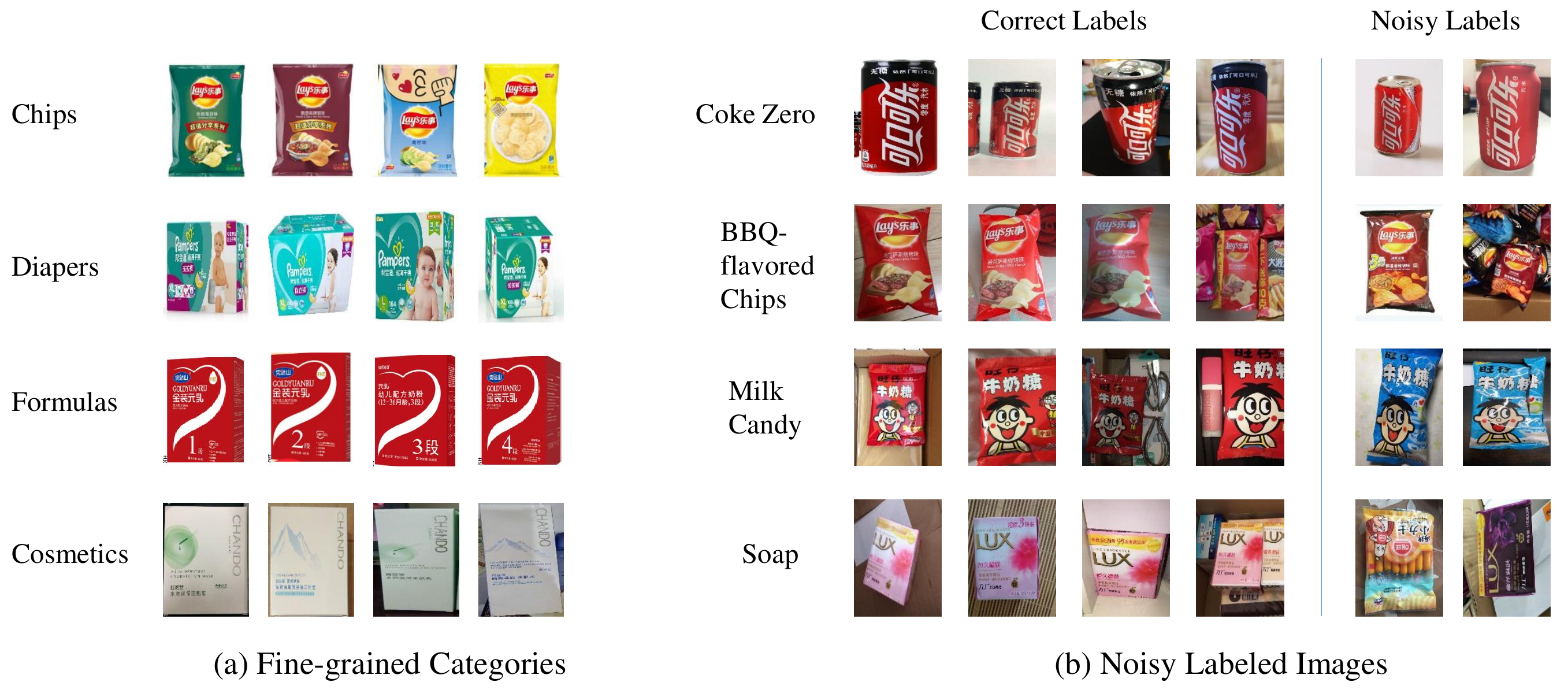}
%\vspace{-0.2cm}
\caption{Image samples of AliProducts, (a) fine-grained categories: from top to bottom are respectively chips in different flavors
, diaper with different sizes, formula with different stages, cosmetic with different functions, and each column is a fine-grained category; (b) noisy labeled images: images from coke zero, bbq-flavored chips, milk candy, soap. As can been see, each category includes massive noisy images.}
%\vspace{-0.5cm}
\end{figure}

\footnotesize\textbf{ImageNet.} ~\cite{imagenet} is an image classification dataset, which contains 1000 classes.
The original dataset has been splitted into 1.28 million training images, 50k validation images.
In this paper, we randomly select 40\% training images for each category and assign them with class label uniformly sampled from the rest categories.
The generate new dataset thus have lot of noises which could be used to evaluate the effectiveness of popular algorithms on noisy image classification.

\footnotesize\textbf{Clothing1M.} ~\cite{cloth1m} is a large-scale clothing dataset, where the images are crawled from online shopping websites. The dataset contains 14 clothes categories and one million noisy labeled images. In addition, 74k manually annotated images are divided into train, validation and test set, with numbers of 50k,14k,10k images, respectively. According to the author's estimate, about 40\% of the labels in the dataset are incorrect because the images has not been fully annotated by human.

\footnotesize\textbf{AliProducts$\footnote{https://tianchi.aliyun.com/competition/entrance/231780/information}$.} is a large-scale noisy and fine-grained product dataset, which includes 50,000 categories.
The images are crawled from image search engine and other web sources by using 50,000 product SKU names. The dataset covers foods, snacks, drinks, cosmetics and other daily products and the categories are in SKU (Stock Keeping Unit) level and specific to flavor, capacity, function or even the batch of the production. Therefore, some of the categories might have great difficulty in visual distinguishing due to the fine-grained attribute.
AliProducts contains 2.5 million training images without any human annotation, consequently contain massive noisy labels, as show in Fig. 3. Totally 148K manually annotated images are used as validation set, and another 250K manually annotated images for testing. In addition, we released side information (e.g., hierarchical relationships between classes) concerning these image data, which could be exploited to learn better representations and models. The main difference between AliProducts and other noisy datasets (e.g., Clothing 1M and WebVision) is that AliProducts contains massive fine-grained and real-world noisy images, which is relatively difficult for robust DNN methods to improve.

%画2幅图，增加一个train loss/test loss下降曲线图
\setlength{\tabcolsep}{4pt}
\begin{table}
\begin{center}
\caption{Top1/Top5 accuracy of three different models with ResNext-101
architecture on validation set of WebVision.}
\label{table:headings}
\begin{tabular}{c|c|c|c}
%\toprule  %添加表格头部粗线
\hline
Method & Model-A & Model-B & Model-C \\
\hline
Top1 & 51.05\% & 47.81\% & 55.57\% \\
\hline
Top5 & 74.94\% & 72.08\% & 78.34\% \\
\hline
%\bottomrule  %添加表格头部粗线
\end{tabular}
\end{center}
\smallskip
\end{table}

\subsection{Experiments on WebVision 2.0}
In this subsection, we conduct extensive experiments on WebVision 2.0 dataset to evaluate and demonstrate the effectiveness of proposed SINet. All experiments are implemented using ResNext-101 backbone if there is no special instructions.

\footnotesize\textbf{Training Strategy and Comparison.}
We conduct three training strategies with a standard ResNext-101 architecture, resulting in three models, which are described as follow.

\footnotesize\textbf{Model-A.} The model was trained by directly using all the training data.

\footnotesize\textbf{Model-B.} The model was trained by using the high-confidence images without reweighting in training loss.

\footnotesize\textbf{Model-C.} The model was trained with proposed training strategy, where the confidence score is multiplied  on the loss of corresponding image for reweighting.

%
%%画2幅图，增加一个train loss/test loss下降曲线图
%\setlength{\tabcolsep}{4pt}
%\begin{table}
%\begin{center}
%\caption{Top-1 and Top-5 accuracy of three different models with ResNext-101
%architecture on validation set of WebVision.}
%\label{table:headings}
%\begin{tabular}{c|c|c}
%%\toprule  %添加表格头部粗线
%\hline
%Method & Top1 & Top5 \\
%\hline
%Model-A & 51.05\% & 74.94\% \\
%\hline
%Model-B & 47.81\% & 72.08\% \\
%\hline
%Model-C & 55.57\% & 78.34\%  \\
%\hline
%%\bottomrule  %添加表格头部粗线
%\end{tabular}
%\end{center}
%\vspace{-0.2cm}
%\end{table}
%\setlength{\tabcolsep}{1.4pt}

The top1/top5 results of three models on the validation set of WebVision are reported in Table 1. The result shows Model-A with all training data significantly outperforms the Model-B with subset of clean data, with improvements of 3.24\%/2.86\% for top1/top5 accuracy.
This is due to that it is hard to distinguish all the clean labeled samples from those images with heavy noises.
In addition, Model-C with our proposed method significantly outperforms Model-A, with improvements of 4.52\%/3.40\% on top1/top5 accuracy.
It is obviously that our proposed method could better explore those clean labels from those noise samples.
These improvements are significant on WebVision Challenge with such a large scale noisy dataset,
which demonstrate the effectiveness of our method.

~\\
\footnotesize\textbf{Class Relation Graph.}
We investigate different ways for constructing Class Relation Graph. (I) Category Name: We use category name, i.e., cat, lion, apple, with BERT model to extract the word embeddings, then use the similarity comparison of word embeddings to construct the Class Relation Graph. (II) Category Description:  We use category descriptions in WordNet , i.e., \emph{Siamese cat: A slender short-haired blue-eyed breed of cat having a pale coat with dark ears paws face and tail tip}, with BERT and LSTM to extract the textual embeddings, then use the similarity comparison of textual embeddings to construct the Class Relation Graph.  (III) Hierachical WordNet: We directly use the prior knowledge of Hierachical WordNet based on the shortest path between two category to establish the Class Relation Graph.
Experimental results of using these three types of class relation graph are shown in Table~\ref{exp_lrg}. Obviously, by introducing these side information, the performance could improve a lot than original Model-A, which shows the effectiveness of proposed class relation graphs. Also, these three types of class relation graph are complementary, and combining of them could also boost the performance.

\begin{minipage}{\textwidth}
\begin{minipage}[t]{0.45\textwidth}
\centering
\makeatletter\def\@captype{table}\makeatother
\setlength{\belowcaptionskip}{10pt}%
\caption{Class Relation Graph construction with different strategies. (I) Category Name (CN) (II)Category Description (CD) (III) Hierachical WordNet (HW)}
\begin{tabular}{c|c|c}
\hline
Strategy & Top1 & Top5 \\
\hline
CN & 54.18\% & 76.84\% \\
%\hline
CD & 55.63\% & 78.15\% \\
%\hline
HW & 55.71\% & 78.42\%  \\
\hline
CN+HW & 55.79\% & 78.46\%  \\
CD+HW & \textbf{55.98\%} & \textbf{78.62}\%  \\
\hline
\end{tabular}
\label{exp_lrg}
\end{minipage}
\quad
\begin{minipage}[t]{0.45\textwidth}
\centering
\makeatletter\def\@captype{table}\makeatother
\setlength{\belowcaptionskip}{10pt}%
\caption{Visual Prototype Generation with different strategies. (I) with Reweighting: Topk matched images with matching score as weighting coefficient (II) without Reweighting: Topk matched images averaged}
\begin{tabular}{c|c|c}
\hline
Strategy & Top1 & Top5 \\
\hline
%\hline
Constant & 55.68\% & 78.45\% \\
Weighting & \textbf{55.98\%} & \textbf{78.62}\%  \\
%\hline
\hline
\end{tabular}
\label{exp_vpg}
\end{minipage}
%\vspace{-0.3cm}
\end{minipage}
\setlength{\tabcolsep}{10pt}

%%画2幅图，增加一个train loss/test loss下降曲线图
%\setlength{\tabcolsep}{4pt}
%\begin{table}
%\begin{center}
%\caption{Class Relation Graph construction with different strategies. (I) Category Name (CN) (II)Category Description (CD) (III) Hierachical WordNet (HW)}
%\label{table:headings}
%\begin{tabular}{c|c|c}
%%\toprule  %添加表格头部粗线
%\hline
%Strategy & Top1 & Top5 \\
%\hline
%CN & 54.18\% & 76.84\% \\
%%\hline
%CD & 55.63\% & 78.15\% \\
%%\hline
%HW & 55.71\% & 78.42\%  \\
%\hline
%CN+HW & 55.79\% & 78.46\%  \\
%CD+HW & \textbf{55.98\%} & \textbf{78.62}\%  \\
%\hline
%%\bottomrule  %添加表格头部粗线
%\end{tabular}
%\end{center}
%\end{table}
%\setlength{\tabcolsep}{1.4pt}

~\\
\footnotesize\textbf{Visual Prototype.}
We investigate different ways for visual prototype generation. (I) Constant: We does not
use any weight operations for the images in Visual Prototype candidates. In this case,
the visual prototype representation in Eq.~(\ref{vk_soft}) is reduced as the
mean of candidates representations.  (II) Weighting: We use the method described in Eq.~(\ref{vk_soft}) as the weighting
operation. Since $p_{i}$ is the importance score, we use the soft weighted representation as the visual prototype.
Experiments of using two types of strategies are shown in Table~\ref{exp_vpg}.
It shows that when paying more attention on top-ranked images, the generated visual prototypes are better then simple average all the feature representations.

~\\
\footnotesize\textbf{Noise Weighting.}
In this section, we conduct ablation analysis on the hyper-parameters
for our proposed noisy weighting method, and
discuss how they affect the recognition performance.
(I) Shift factor $\alpha$: $\alpha$ controls the amount of noisy data actually participating in the model training, the weight of images whose Euclid distance larger than $\alpha$ is 0, that is equivalent to deleting them from the training set, and only use the the images with Euclid distance smaller than $\alpha$ to train the model.
(II) Contrast factor $\beta$: We introduce the contrast
parameter $\beta$ to sharpen the differences of the scores.
Experiments of different $\alpha$ and $\beta$ are shown in Table~\ref{exp_alpha} and~\ref{exp_beta}, respectively.
Typically, $\alpha=1.2$ could keep most of cleaning samples.
$\beta=1.5$ is a proper value to map the score to sampling weight and handle the noise data.

%\begin{figure}[htbp]
%\centering
%\includegraphics[height=4.2cm, width=12cm]{fig/noiseweighting}
%\caption{Image confidence score generated by noise weighting module.}
%\end{figure}
~\\
\begin{minipage}{\textwidth}
\begin{minipage}[t]{0.45\textwidth}
\centering
\makeatletter\def\@captype{table}\makeatother
\setlength{\abovecaptionskip}{0pt}%
\setlength{\belowcaptionskip}{10pt}%
\caption{The effect of Shift factor $\alpha$ on the performance of WebVision validation set.}
\begin{tabular}{c|c|c}
\hline
$\alpha$ & Top1 & Top5 \\
\hline
0.8 & 53.25\% & 76.24\% \\
%\hline
1.0 & 55.43\% & 78.31\% \\
%\hline
1.2 & \textbf{55.98\%} & \textbf{78.62}\%  \\
%\hline
1.4 & 55.76\% & 78.48\%  \\
\hline
\end{tabular}
\label{exp_alpha}
\end{minipage}
\quad
\begin{minipage}[t]{0.45\textwidth}
\centering
\makeatletter\def\@captype{table}\makeatother
\setlength{\abovecaptionskip}{0pt}%
\setlength{\belowcaptionskip}{10pt}%
\caption{The effect of Contrast factor $\beta$ on the performance of WebVision validation set.}
\begin{tabular}{c|c|c}
\hline
$\beta$ & Top1 & Top5 \\
\hline
%\hline
1.0 & 55.68\% & 78.45\% \\
1.5 & \textbf{55.98\%} & \textbf{78.62}\%  \\
2.0 & 55.82\% & 78.56\%  \\
2.5 & 55.35\% & 78.24\%  \\
%\hline
\hline
\end{tabular}
\label{exp_beta}
\end{minipage}
\end{minipage}

~\\
~\\
\footnotesize\textbf{Final results on the WebVision challenge.}
We further evaluate the performance of our proposed SINet with various networks architectures, including ResNext101, SE-ResNext101, SE-Net154. Results are reported in Table~\ref{exp_all}. As can be found, SE-Net154 substantially performs ResNext101 and SE-ResNext101 on WebVision validation set, with top1/top5 improvements of 1.31\%/1.46\% and 1.30\%/1.27\%, while
SE-ResNext101 and ResNext101 has similar performance with a marginal performance gain obtained.
Our final results were obtained with ensemble of five models. We had the
best performance at a Top 5 accuracy of 82.54\% on the WebVision challenge 2019.
%It outperforms the 2nd one by a margin of about 0.5\%. Although the 82.54\% top5 accuracy is not comparable to human performance on the ImageNet, we achieve this result by using weakly-supervised noisy data without any manual annotation.

\setlength{\tabcolsep}{4pt}
\begin{table}
\begin{center}
\caption{Performance of SINet with various networks on WebVision validation set.}
\label{exp_all}
\begin{tabular}{c|c|c|c}
%\toprule  %添加表格头部粗线
\hline
Method & ResNext101 & SE-ResNext101 & SENet154  \\
\hline
Top1 & 55.56\%  &55.57\% & 56.87\%   \\
\hline
Top5 & 78.15\%  &78.34\% & 79.61\%   \\
\hline
%\bottomrule  %添加表格头部粗线
\end{tabular}
\end{center}
%\vspace{-0.5cm}
\end{table}

\subsection{Comparisons with the State-of-the-art Methods}
%AAAAAAAAAAAAAAAAAAAAAA
%\subsubsection{Implementation Details.} dataset
%\subsubsection{Large Scale Web Noisy Data.} dataset
%\subsubsection{Final results on the WebVision challenge.} dataset
To further explore the effectiveness of our proposed SINet, we conduct extensive comparisons with recent state-of-the-art approaches developed specifically for learning from noisy labels, such as CleanNet, MetaCleaner and MentorNet. For fairness, our comparisons are based on the same
CNN backbone, i.e., ResNet50.

\footnotesize\textbf{WebVision1.0 and ImageNet.}
We evaluate our SINet on ImageNet, by adding 40\% noise ratio with uniform flip. The top-1 accuracy is 66.47/69.12 for ResNet50 without/with SINet.
It further shows the power of SINet for large-scale noisy image recognition.
By following~\cite{ref18}, we use the training set of WebVision1.0 to train the models, and test on the validation sets of WebVision1.0 and ImageNet. Both of them has same 1000 categories. Full results are presented in Table~\ref{exp_webimagenet}. SINet improves the performance of our baseline significantly, and our results compare favorably against recent CurriculumNet, CleanNet and MentorNet with consistent improvements.
%ranged from about 1.5\% to 3.3\%.

\setlength{\tabcolsep}{4pt}
\begin{table}
\begin{center}
\caption{Comparisons on Webvision1.0 and ImageNet. The models are trained on WebVision1.0 training set and tested on WebVision1.0 and ImageNet validation sets.}
\label{exp_webimagenet}
\begin{tabular}{c|c|c}
%\toprule  %添加表格头部粗线
\hline
\multirow{2}*{Method} & WebVision1.0 & ImageNet \\
%\hline
%Method & {WebVision1.0} & \multicolumn{2}{|c|}{WebVision2.0}  & ImageNet \\
\cline{2-3}
~ & Top1/Top5 & Top1/Top5 \\
\hline
Baseline~\cite{ref12} & 67.8(85.8) & 58.9(79.8) \\
\hline
CleanNet~\cite{ref12} & 70.3(87.8) & 63.4(84.6) \\
\hline
MentorNet~\cite{ref11} & 70.8(88.0) & 62.5(83.0) \\
\hline
CurriculumNet~\cite{ref18} & 72.1(89.2) & 64.8(84.9) \\
\hline
\hline
Our Baseline & 69.9(87.4) & 63.2(83.8) \\
\hline
SINet & \textbf{73.8(90.6)} & \textbf{66.8(85.9)} \\
\hline
%\bottomrule  %添加表格头部粗线
\end{tabular}
\end{center}
%\vspace{-0.5cm}
\end{table}

%%\footnotesize\textbf{WebVision1.0.}  We also validate the proposed method on WebVision 1.0, which contains 1000 categories and 2.4 million images.
%%Comparisons with state-of-the-art methods are shown in Table~\ref{exp_webimagenet}.
%
%\setlength{\tabcolsep}{1.4pt}
%\setlength{\tabcolsep}{4pt}
%\newcommand{\tabincell}[2]{\begin{tabular}{@{}#1@{}}#2\end{tabular}}
%\begin{table}
%\begin{center}
%\caption{Experimental results on Clothing1M.
%Clean set is used in CleanNet~\cite{ref12} to obtain the validation set.
%To keep same data setting, we use the 25k clean images to construct the visual prototype and use 1M noisy
%training set with confidence scores to train our SINet, and then fine-tune it on 25k clean images. Furthermore,
%we achieve the state-of-the-art performance on the setting
%of Noise1M+Clean(50k), which illustrate the  robustness of our SINet on noisy label recognition.}
%\label{exp_cloth1m}
%\begin{tabular}{c|c|c|c|c|c|c}
%%\toprule  %添加表格头部粗线
%\hline
%\multirow{2}*{\tabincell{c}{Noise1M\\+Clean(25k)}} & Method & CleanNet($w_{hard}$) & CleanNet($w_{soft}$) & MetaCleaner & DeepSelf & Ours \\
%\cline{2-7}
%~ & Accuray & 74.15 &  74.69 & 76.00 & 76.44 & \textbf{77.26} \\
%\hline
%\hline
%\multirow{2}*{\tabincell{c}{Noise1M\\+Clean(50k)}} & Method & LossCorrect & CleanNet($w_{soft}$) &  MetaCleaner & DeepSelf & Ours \\
%\cline{2-7}
%~ & Accuray & 80.38 &  79.9 & 80.78 &  81.16 & \textbf{81.32} \\
%\hline
%%\bottomrule  %添加表格头部粗线
%\end{tabular}
%\end{center}
%%\vspace{-0.5cm}
%\end{table}

\setlength{\tabcolsep}{1.4pt}
\setlength{\tabcolsep}{4pt}
\newcommand{\tabincell}[2]{\begin{tabular}{@{}#1@{}}#2\end{tabular}}
\begin{table}
\begin{center}
\caption{Experimental results on Clothing1M.
Clean set is used in CleanNet~\cite{ref12} to obtain the validation set.
To keep same data setting, we use the 25k clean images to construct the visual prototype and use 1M noisy
training set with confidence scores to train our SINet, and then fine-tune it on 25k clean images. Furthermore,
we achieve the state-of-the-art performance on the setting
of Noise1M+Clean(50k), which illustrate the  robustness of our SINet on noisy label recognition.}
\label{exp_cloth1m}
\begin{tabular}{c|c|c|c|c|c}
%\toprule  %添加表格头部粗线
\hline
\multirow{2}*{\tabincell{c}{Noise1M\\+Clean(25k)}} & Method & CleanNet~\cite{ref12} & MetaCleaner~\cite{ref23} & DeepSelf~\cite{ref22} & Ours \\
\cline{2-6}
~ & Accuray & 74.69 & 76.00 & 76.44 & \textbf{77.26} \\
\hline
\hline
\multirow{2}*{\tabincell{c}{Noise1M\\+Clean(50k)}} & Method & CleanNet~\cite{ref12} &  MetaCleaner~\cite{ref23} & DeepSelf~\cite{ref22} & Ours \\
\cline{2-6}
~ & Accuray & 79.9 & 80.78 &  81.16 & \textbf{81.32} \\
\hline
%\bottomrule  %添加表格头部粗线
\end{tabular}
\end{center}
%\vspace{-0.5cm}
\end{table}

\footnotesize\textbf{Clothing1M.}
For Clothing1M, we consider the state of the art results in~\cite{ref23}, which use both noisy and clean set to train the model.
Following~\cite{ref23}, we conduct two experiments.
First we use the 25k clean set to construct the visual prototype and apply noise weighting to one million noisy data, and then
use the images with confidence score to train the model. Second, we conduct the same experiment, but with all the clean training set (50k). As shown in Table~\ref{exp_cloth1m}, our SINet outperforms CleanNet, MetaCleaner and DeepSelf, which demonstrates its effectiveness.

\footnotesize\textbf{AliProducts.}
The existed benchmarks with noisy labels are relatively small in the scale of categories or images. To further explore the effectiveness of SINet, we conduct experiments on our released AliProducts, which is a large-scale product dataset with noisy labels and the hierarchical category relations is also provided.
We use the hierarchical category relations to construct the class relation graph, and then combine it with images to construct the visual prototype.
Finally, we use the images with confidence scores to train the model. As shown in Table~\ref{ exp_AliProducts}, our SINet outperforms
all other approaches, which illustrates
that SINet is more robust to noisy labels.

\setlength{\tabcolsep}{1.4pt}
\setlength{\tabcolsep}{4pt}
\begin{table}
\begin{center}
\caption{Comparison with the State-of-The-Art on AliProducts dataset.}
\label{ exp_AliProducts}
\begin{tabular}{c|c|c|c|c|c}
%\toprule  %添加表格头部粗线
\hline
Method & Baseline & CurriculumNet~\cite{ref18} & CleanNet~\cite{ref12} & MetaCleaner~\cite{ref23} & Ours \\
\hline
Accuray & 85.35\% & 85.69\% & 86.13\%  & 85.92\% &  \textbf{86.29\%} \\
\hline
%\bottomrule  %添加表格头部粗线
\end{tabular}
\end{center}
%\vspace{-1cm}
\end{table}

\section{Conclusions}
In this paper, we presented a novel method, which can learn to generate a visual prototype for each category, for training deep
CNNs with large-scale real-world noisy labels. It mainly consists of two submodules. The first module, Visual Prototype can generate a clean representation from the noisy images for every category by integrate the noisy images with side information. The second module, namely Noise Weighting, can estimate the confidence scores of all the noisy images and rank images with confidence scores by analyzing their deep features and Visual Prototype. Via SINet, we can train a high-performance CNN model, where the negative impact of noisy labels can be reduced substantially. We conduct extensive experiments on WebVision, ImageNet, Clothing1M, as well as collected AliProducts, where it achieves state-of-the-art performance on all benchmarks. Future work could aim to train an end-to-end
DNNs with Side Information to handle the noisy label recognition.

\bibliographystyle{splncs04}
\bibliography{egbib}
\end{document}